**Deep Kernel Methods Learn Better: From Cards to Process Optimization**

Mani Valleti,[1,a] Rama K. Vasudevan,[2] Maxim A. Ziatdinov,[2, 3] and Sergei V. Kalinin[4,b]

[1] Bredesen Center for Interdisciplinary Research, University of Tennessee, Knoxville, TN 37916 USA

[2] Center for Nanophase Materials Sciences, Oak Ridge National Laboratory, Oak Ridge, TN 37831 USA

[3] Computational Sciences and Engineering Division, Oak Ridge National Laboratory, Oak Ridge, TN 37831 USA

[4] Department of Materials Science and Engineering, University of Tennessee, Knoxville, TN 37916 USA

The ability of deep learning methods to perform classification and regression tasks relies heavily on their capacity to uncover manifolds in high-dimensional data spaces and project them into low-dimensional representation spaces. In this study, we investigate the structure and character of the manifolds generated by classical variational autoencoder (VAE) approaches and deep kernel learning (DKL). In the former case, the structure of the latent space is determined by the properties of the input data alone, while in the latter, the latent manifold forms as a result of an active learning process that balances the data distribution and target functionalities. We show that DKL with active learning can produce a more compact and smooth latent space which is more conducive to optimization compared to previously reported methods, such as the VAE. We demonstrate this behavior using a simple cards data set and extend it to the optimization of domain-generated trajectories in physical systems. Our findings suggest that latent manifolds constructed through active learning have a more beneficial structure for optimization problems, especially in feature-rich target-poor scenarios that are common in domain sciences, such as materials synthesis, energy storage, and molecular discovery. The jupyter notebooks that encapsulate the complete analysis accompany the article.

[a] svalleti@vols.utk.edu
[b] sergei2@utk.edu

The ability of machine learning to perform tasks such as semantic image segmentation,[1-3] regression,[4, 5] classification,[6, 7] and unsupervised learning[8, 9] has made it a powerful tool in a wide range of applications. ML offers several advantages, including the ability to automatically learn patterns and relationships from data, process large amounts of information quickly and accurately, and make predictions or decisions based on the learned information. Additionally, these techniques can adapt to changing data and learn from new examples, making them a useful tool for both exploratory data analysis and predictive modeling.

Over the last several years, active learning methods have proven to be highly effective in domain-specific tasks such as robotics[10-13] and autonomous driving.[14-18] and extended to areas such as materials and molecular discovery,[19-24] optimization of processes ranging from material synthesis[25-27] to battery charging.[28-30] For materials synthesis, this can include the variation of the processing temperature during annealing or substrate temperature and oxygen pressure during film growth. For batteries, this includes optimal voltage- and current charging profiles. For molecular systems, the relevant problem is the exploration of high-dimensional chemical or chemical fingerprint space towards target functionality.

The ability of machine learning (ML) algorithms, both supervised and unsupervised, to effectively analyze high-dimensional datasets depends on their ability to discover and represent low-dimensional manifolds that capture the underlying behavior of the data. One such method for discovering these manifolds is the variational autoencoder (VAE), which learns representations of the data by encoding it into a lower-dimensional latent space with one neural network and then reconstructing it back from the latent space with another neural network. The VAE has pronounced ability to disentangle these representations, with the latent variables representing the intrinsic factors of variation in the dataset. However, the relationship between these latent variables and the underlying mechanisms that generate the data is not yet fully understood, despite extensive empirical demonstrations. Furthermore, the structure and character of the latent manifolds can be expected to have a significant impact on their performance, and there is still much to be explored and understood in this area.[31]

To address this, various modifications of the original VAE framework have been introduced. One such approach is conditioning the VAEs on the known factors of variation.[32] Alternatively, the VAE network's structure can be modified to separate physical factors of variations from the geometric ones such as rotation and shift.[33] Similarly, the topology of the latent space can be controlled. However, in all these applications, the structure of the latent space and the latent space distributions are determined solely by the input data.

Here we investigate the evolution of the latent spaces in deep kernel learning (DKL) in both static and active settings and apply this approach to a toy card dataset as well as process optimization in high-dimensional trajectory spaces. Unlike VAEs, where the latent space structure is determined solely by the data, DKL constructs the latent space while predicting the target function in the latent space, allowing for the target function to shape the latent manifold. We hypothesize and demonstrate that for the model system, the latent manifold formed during active DKL is more conducive to process optimization than that of the VAE case, with regions corresponding to the optima of the desired behavior occupying a significant portion of the latent space. In contrast, distributions of counterfactual measures are much rougher. These observations suggest that active DKL may have significant implications for optimization problems over large-dimensional non-differentiable spaces, such as molecular property optimization.

## 1. Latent distributions in VAE and rVAE for model data

Variational autoencoders (VAEs) are a class of generative models that learn a low-dimensional representation of high-dimensional data, such as images or time series.[34-39] The idea behind VAEs is to learn a compact and interpretable latent space that captures the essential features of the input data, while also being able to generate new samples that are similar to the training data. VAEs are based on the idea of autoencoders, which consist of an encoder that maps the input data to a lower-dimensional latent space, and a decoder that maps the latent space back to the original data space.

In VAEs, the encoder and decoder are trained to minimize a reconstruction loss that measures the difference between the input data and the reconstructed output. However, to avoid overfitting and encourage a more structured latent space, VAEs also introduce a regularization term that encourages the encodings in the latent space to follow a specific prior distribution, such as a standard normal distribution. This regularization prevents the latent space from overfitting to the training data and helps to produce a smooth and continuous latent space that captures the essential variations in the input dataset, which allows the VAE to interpolate between different data points and generate novel samples. Specifically, the regularization involves minimizing the Kullback-Leibler (KL) divergence between the learned latent distribution and the prior distribution. By minimizing the KL divergence, the VAEs can learn to encode the essential features of the input dataset in the latent space while maintaining the desirable properties of a regularized distribution, such as smoothness and continuity. Therefore, the latent space of the VAEs provides a useful and interpretable representation of the input data that can be used for various downstream tasks. Additionally, VAEs have generative capabilities, which means that they can generate new samples by sampling from the latent space and decoding the samples into the original data space. This makes VAEs a useful tool for various applications in machine learning, such as image generation,[40, 41] noise cleaning,[42, 43] and data compression.[44, 45]

Rotationally invariant autoencoder (rVAE) is a variant of VAE that introduces additional latent dimensions to explicitly capture rotational and/or translational symmetries in the input data.[33, 46-49] By assigning 1-3 latent dimensions to encode the angles or shifts in the input data, rVAE can learn a latent space that is invariant to rotations and/or translations, which can be useful for a variety of scientific applications. For example, in the field of condensed matter physics, rVAE can be used to analyze and compare images of crystal lattices with different orientations, while in microscopy, it can help to improve the accuracy of image segmentation and tracking of moving objects.[46, 48] In materials science, rVAE can aid in the identification and classification of microstructures in materials, where the microstructures may be rotated or shifted with respect to each other.[47, 49] When using rVAE to learn a latent space that is invariant to rotations and/or translations, images that differ only by rotations or translations will have different encodings in the dimensions that explicitly capture these transformations, while sharing the same or similar encodings in the other dimensions. This ensures that the learned model can recognize and differentiate between images based on their orientation or position, while also being able to generalize to new images that may have different orientations or positions.

To illustrate the evolution and formation of the latent distributions in VAEs and rVAEs and the role of prior knowledge in the formation of manifold, we use a toy dataset called the cards dataset. The chosen toy dataset comprises of monochrome images of playing cards from all four suits - clubs, spades, diamonds, and hearts. Despite its apparent simplicity, the playing cards dataset presents several challenges for machine learning algorithms due to its complex symmetries. For instance, the clubs suit has a three-fold symmetry (without the tail) and a mirror symmetry. This means a rotation of 120° each will map a club to itself, while a mirror reflection about any axis will also produce the same club. Similarly, the spades suit also has a mirror symmetry and at high deformations can be difficult to differentiate from the clubs suit. Moreover, horizontally flipped hearts and spades can be nearly identical, with subtle differences that can be challenging to distinguish. Additionally, diamonds have a unique shape that is distinct from the other three suits and rotating them by 90° can lead to degeneracies in the outcomes of applying affine transformations to the dataset. These complexities and symmetries in the playing cards dataset can pose challenges for machine learning algorithms.

In order to analyze the capabilities of VAEs and rVAEs and introduce traceable variability to the cards dataset in addition to the class labels, we apply affine transformations to the images. We randomly rotate and shear each image independently, with values chosen from ranges of [-120°, 120°] and [-20°, 20°], respectively. The limits of these transformations are chosen such that the images are highly distorted, and the rotations will bring the rotational symmetries of the all the four suits into play. The images are sheared equally in $x$ and $y$ directions. We select 2,000 images from each suit to ensure an equal representation of all classes in the dataset and apply the affine transformations to each image to produce a final dataset of 8,000 images for our analyses. By saving the ground truth values separately, we can analyze how VAEs and rVAEs distribute these variabilities in the latent space and visualize complex symmetries and deformations of the playing cards. The latent dimensions $z_1$, $z_2$ will be referred to as the regular latent dimensions for both VAE and the rVAE and the latent dimension that explicitly captures the angle in the case of rVAE will be referred to as the angular dimension ($\theta$). The number of regular latent dimensions ($z_1$, $z_2$) for the cards dataset case is set to 2 for both VAE and the rVAE cases. However, it should be noted that the rVAE has a third latent angular dimension ($\theta$) to explicitly capture the rotations.

The construction of VAE's latent space can be visualized using two different methods, and one being the decoded latent space, or latent representation. This method discards the input dataset after the training and makes use of only the decoder or generator part of the VAE to form the representation, which can be visualized and explored to gain insight into the underlying structure of the data. This method involves sampling the regular latent space ($z_1$, $z_2$) uniformly and decoding the samples back into the image space while holding any additional latent dimensions (such as the angular dimension in the case of rVAE) at constant values, usually zero. The prior distribution for all the regular latent dimensions is set to a unit normal. Hence, the limits for the regular dimensions for sampling in this method are set to [-1.5, 1.5] for each latent dimension. The decoded images are then plotted in a grid that corresponds to their positions in the latent space, allowing us to observe the continuous transformations of the images that correspond to the changes in the latent space. Figure 1a and 1e show the decoded latent spaces for VAE and rVAE, respectively. In the decoded latent space of VAE (Figure 1a), images with multiple orientations are observed because the VAE treats images with different orientations as separate entities. The angular encoding for all the images in the decoded latent space of rVAE (Figure 1e) is set to zero, leading to orientations due to the baseline zero of the encodings for each class. In complex datasets, datapoints from a single class are seldom identified with multiple baseline zeros, as evident in the case of hearts suit in Figure 1e, where hearts with multiple orientations are found in the decoded latent space. Furthermore, the decoded latent spaces provide insights into how the image from one class gradually transforms into the image from a different class in the latent space. This process leads to the formation of images in the decoded latent space that are not only part of the input dataset but can also be non-physical.

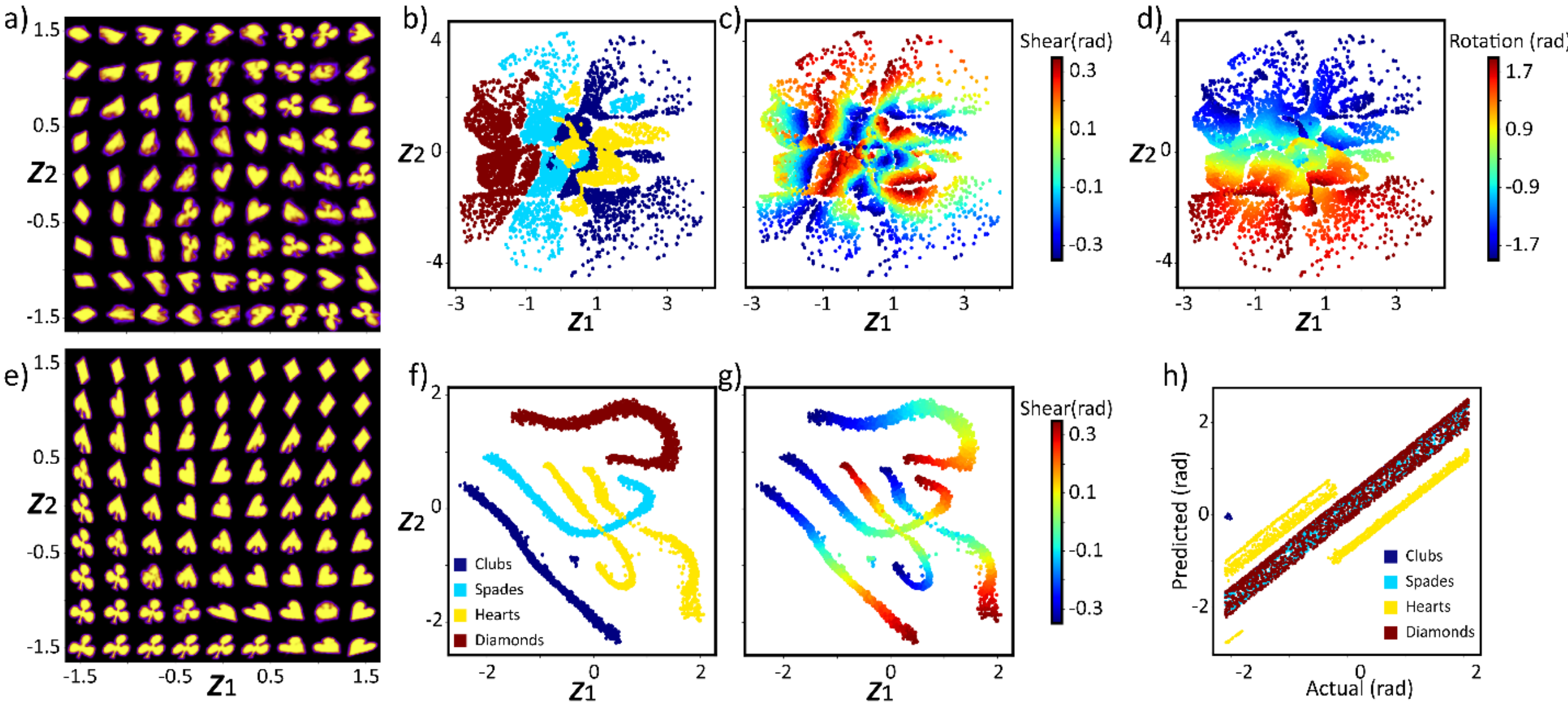

**Figure 1.** Decoded latent space of (a) VAE, (e) rVAE. Latent distributions of VAE colored using (b) class label, (c) ground truth shear value in radians, and (d) ground angle of rotation in radians. Latent distributions of rVAE colored using (f) ground truth class labels and (g) ground truth shear value in radians. (h) Ground truth angle of rotation *vs*. the angle encoded by the angular latent dimension of rVAE in radians. The color bar for (b) is the same as the one used in (f) and (h).

The second method of visualizing the latent space is through latent distributions. Unlike the decoded latent space, which uses the decoder to generate samples from the latent space, latent distributions focus on the latent variables of the input dataset embedded by the encoder. By examining the distributions of these latent variables, researchers can gain insights into the characteristics of the data, the properties of the latent space, and the behavior of the VAE.

Additionally, by using the latent distributions to model the data, researchers can generate new samples that preserve the underlying patterns and structures of the original data. This method involves plotting the input dataset's images as scatter points in the latent space, with the option to color the points using ground truth class labels or features of variability such as class labels, angles of rotation, and shear. By examining the distribution of the scatter points, we can identify the regions of the latent space that correspond to each class of the dataset, providing insights into the construction of the latent space. This method can also help us visualize the variation of continuous ground truth entities in the latent space and identify gaps or inconsistencies that may affect the VAEs' performance. Figures 1b-d show the latent distributions of VAE, with the scatter points colored using ground truth class labels (Figure 1d), shear (Figure 1c), and rotations (Figure 1d). The rVAE counterparts are shown in Figure 1f for class labels and Figure 1g for ground truth shear values. The distribution of rotations for rVAE is omitted intentionally as it is explicitly encoded in the angular dimension, resulting in a noisy distribution in the latent space of $z_1$-$z_2$. The VAE's latent distributions exhibit a higher degree of entanglement than those of the rVAE, likely due to the difficulty of representing three factors of variation (ground truth class label, shear, and rotations) in a two-dimensional space. This is reflected in the higher mutual information between latent dimensions and the less well-defined regions of the latent space associated with each class. As a result, complex structures such as overlapping of images from different classes (Figure 1b) and nonlinear (albeit smooth) variation of continuous ground truth variabilities (Figures 1c and 1d) form in the VAE's latent space. Conversely, the rVAE disentangles the classes and the latent distribution has well-defined spaces associated with each class (Figure 1f) and a smooth linear variation of ground truth shear values (Figure 1g). Only the hearts suit in the rVAE's case appears to be encoded in two disconnected branches and overlaps with the spades suit (Figure 1f). The shear distribution in the rVAE's latent space (Figure 1g) shows a linear distribution of ground truth values even in the case of disconnected hearts' branches.

Figure 1h displays the relationship between the encoded angles in the angular dimension ($\theta$) of the rVAE and the ground truth rotations, with datapoints colored according to their ground truth class labels. The presence of multiple branches for the hearts suit indicates the encoding of multiple baseline zeros for this class. The standard deviation of the encodings of a given ground truth rotation can be attributed to the dataset's complexity. Since the rVAE performs unsupervised learning of rotations, the high shear present in the inputs poses a challenge for the rVAE to achieve perfect encoding of the ground truth rotations. The complete analysis of the cards dataset using multiple variations of VAE is available in our review article on VAEs. [Reference to be added]

Deep learning models such as VAEs have been used in high-dimensional process optimization problems which optimize a scalar target function in the low-dimensional latent space.[50-53] However, their latent space has limitations when it comes to modeling categorical and continuous target functions. The VAEs learn a low-dimensional representation of high-dimensional data by minimizing the reconstruction error between the input and the reconstructed input, while regularizing the latent space to have a unit normal distribution. However, this regularization only ensures that the latent space is smooth along the input dimensions and not necessarily smooth in terms of the target function. This can result in a complicated and non-smooth latent space, making it difficult to optimize the target function using descent-based methods (Figure 1c). In some cases, the target function may have a non-differentiable relationship with the input space. This complex relationship between the latent space and the target function can lead to non-differentiable distributions of the target function in the latent space, which can further exacerbate this problem. This is especially problematic in process optimization problems, as the search algorithms rely heavily on the differentiability of the target function for efficient optimization. Even when the target function has a differentiable relationship with the input space, oftentimes it results is a complex and cluttered latent space that is challenging to navigate for the optimization algorithms. Furthermore, descent-based optimization methods such as stochastic gradient descent (SGD) tend to get stuck in local minima in the latent space, resulting in suboptimal solutions. On the other hand, Bayesian optimization (BO) requires an extensive number of points to be sampled to explore a sufficient region of the latent space, making it computationally expensive. In the case of the rVAE, we found that it works well only when one of the variabilities in the dataset is rotations, and the remaining variabilities are not complicated. In other cases where multiple variability factors are involved, such as ground truth class labels and shear, the latent space becomes complex and challenging to navigate for the optimization algorithms. BO remains an effective search method in both cases for differentiable target function distributions, but it requires a considerable number of function evaluations to find the optimal solution.

## 2. Latent distributions for the DKL of model data

Deep Kernel Learning (DKL) poses to be one of the potential solutions to the problems that are discussed so far. Deep Kernel Learning (DKL) is a method that combines the power of deep neural networks with the flexibility of Gaussian Processes (GPs) to model and make predictions on complex target functions in high-dimensional spaces.[54-58] DKL is able to efficiently explore the high-dimensional input space and learn the relationship between the inputs and target function, making it a powerful tool for optimization and prediction tasks. GP provides a way to model complex functions that are difficult to express in closed form.[59] It works by modeling the target function as a distribution over functions in the input space. In GP, each point in the input space (latent space of DKL) is associated with a random variable that represents the value of the function at that point. Given a set of input data $X = \{x_1, x_2, \ldots, x_N\}$ and their corresponding outputs/targets $y = \{y_1, y_2, \ldots, y_3\}$ the GP models the relationship between the input space and the target space using equation 1, where $K$ is a covariate function (kernel) with hyperparameters $\gamma$.

$$y \sim MutivariateNormal\big(0, K(X|\gamma)\big) \quad (1)$$

For the analyses discussed from here on, we use radial basis function (RBF) kernel with weakly informative priors. The RBF kernel is shown in equation 2, where the hyperparameter $\alpha$ is the variance/amplitude parameter which determines the overall scaling of the kernel. The hyperparameter $l$ is the length scale parameter that controls the smoothness of the function in the input space. $\delta$ and $\sigma_{noise}$ correspond to the Dirac-delta function and the noise term in the regression setting. We use weakly informative priors for the hyperparameters $\alpha$ and $l$, which are given by equations 3.

$$K_{ij} = \alpha \, exp\left( {-0.5\left\| x_i - x_j \right\|^2}/{l^2} \right) + \delta_{ij}\sigma_{noise} \tag{2}$$

$$\alpha \sim logNormal(0,1)$$
$$l \sim logNormal(0,1) \tag{3}$$

The GP posterior prediction is computed analytically and can then be directly sampled to give the predictions on the unseen inputs. Specifically, given the kernel hyperparameters $\gamma = \{\alpha, l\}$, the posterior prediction on the new inputs $\ddot{X}$ is defined by equations 4,

$$p(\ddot{y}|\ddot{X}, y, X) = N(\ddot{\mu}, \ddot{\Sigma}) \tag{4}$$
$$\ddot{\mu} = K(\ddot{X}, X|\gamma)K(X|\gamma)^{-1}y$$
$$\ddot{\Sigma} = K(\ddot{X}|\gamma) - K(\ddot{X}, X|\gamma)K(X|\gamma)^{-1}K(X, \ddot{X}|\gamma)$$

where $K(X|\gamma)$ and $K(\ddot{X}|\gamma)$ are the matrices obtained by computing kernel function (equation 2) between all pairs of input data ($X$) and all pairs of unseen data ($\ddot{X}$) respectively. $K(\ddot{X}, X|\gamma)$ is obtained by evaluating the kernel function between the input data and the unseen data. GP is characterized by a prior mean function (in our case assumed to be a constant zero-function) and a prior covariance function ($K_{ij}$), which capture the prior assumptions about the behavior of the function. The mean function specifies the average value of the function, while the covariance function captures the degree of similarity between two points in the input space. The smoothness of the function is controlled by the covariance function, which ensures that nearby points in the input space have similar function values. The kernel hyperparameters are usually inferred using either a Markov chain Monte Carlo (MCMC) sampling or variational inference approaches.[60]

One of the major limitations of GP is unlike deep learning, the GP does not learn representations from the data. On the other hand, DKL combines the deep neural networks that learn low-dimensional representations from a high dimensional input data with GP. In DKL, the high-dimensional input data is projected into the low-dimensional latent space and the GP models the targets as a function of the latent space. This structure of neural network+GP is referred to as 'deep-kernel' and is given by equation 5,

$$K_{DKL}(X|\omega, \gamma) = K_{base}(g(X|\omega)|\gamma) \tag{5}$$

where $g(X|\omega)$ are the latent representations of the input dataset $X$. The weights of the neural network $g$ that projects the high-dimensional inputs into a low-dimensional latent space are represented by $\omega$ and $K_{base}$ is the GP-kernel (an RBF kernel as discussed in equation 2). The hyperparameters of the deep-kernel are the weights of the neural network ($\omega$) in conjunction with the hyperparameters of GP ($\gamma$). These hyperparameters can be deterministic and optimized with respect to marginal likelihood of the data using stochastic gradient descent. This method tends to overfit the data albeit being computationally cheap. The Bayesian way is to represent the weights of the neural network ($\omega$) by probabilistic distributions. The posterior distributions of hyperparameters of the deep-kernel ($\gamma$ and $\omega$) are then learned in conjunction using the variational inference approximation. However, it should be noted that the posteriors of the hyperparameters can be sampled using MCMC but avoided here due to computational inefficiencies.

In summary, the DKL employs a two-component process to model the target function ($y$) as a function of the input space ($X$). The first component, which we will refer to as the encoder of the DKL, a neural network maps the inputs ($X$) to a lower-dimensional latent space ($D \equiv g(X|\omega)$). Subsequently, the second component, a GP layer then predicts the distributions of the targets in the latent space ($D$). The schematic of the DKL is shown in Figure 2. The latent space acts as the input space to the GP, while the values of the target function act as outputs. The latent space of DKL stands in quite contrast to VAE where the latent space of former is not regularized. The latent space of DKL is not explicitly regularized but is rather structured in a way that the output distribution is smooth, which makes it well-suited to be modeled by a Gaussian Process (GP). In contrast, the latent space of VAEs is regularized to follow a specific prior distribution such as a Gaussian distribution, which encourages the latent space to be smooth and continuous. However, the DKL's approach results in a more compact and differentiable distribution of targets in the latent space compared to VAEs, making it more suitable for optimization problems over high-dimensional spaces. Additionally, by optimizing the latent space and the hyperparameters jointly, the DKL can learn a more flexible representation of the data that can capture complex patterns in the data and generalize well to new data.

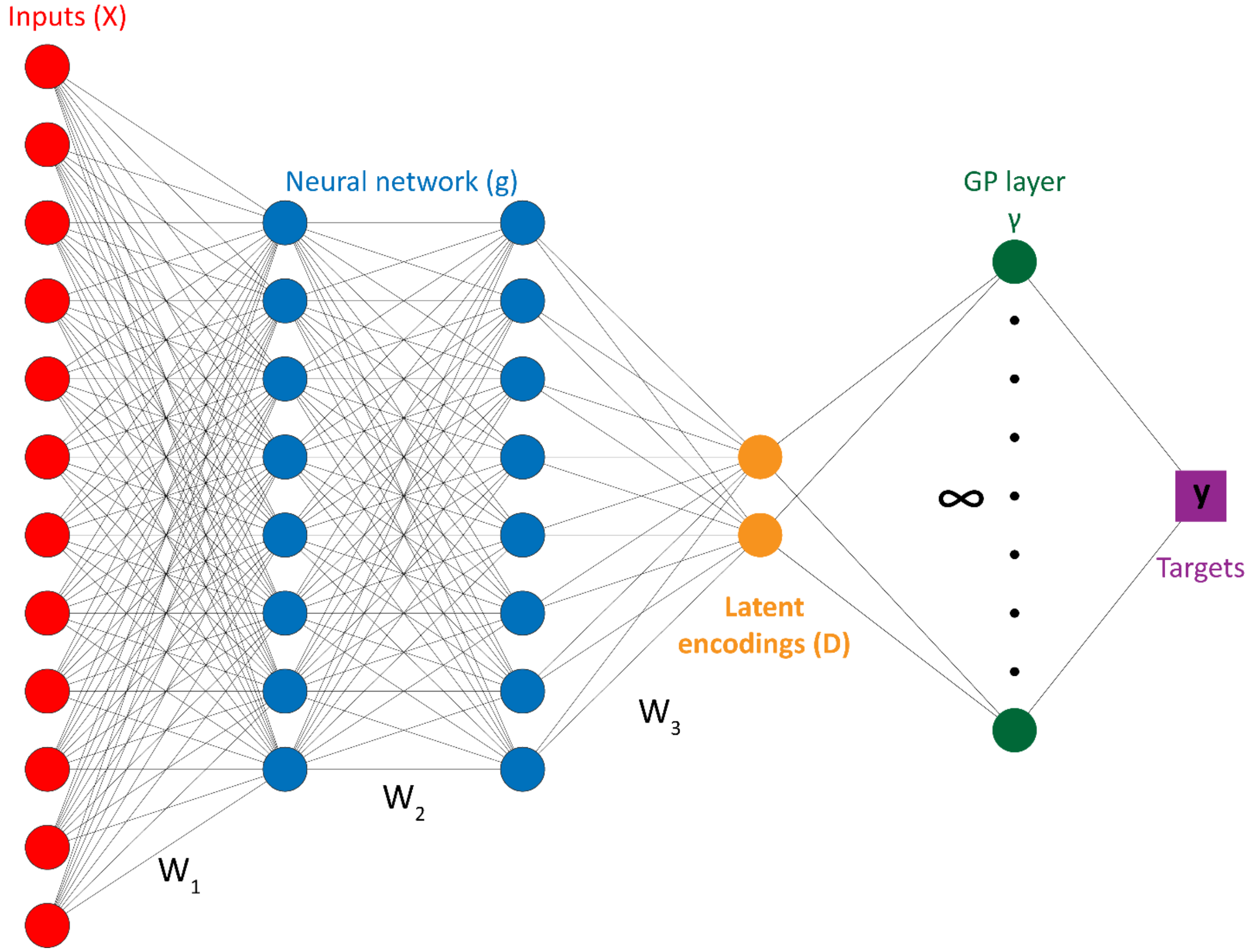

**Figure 2.** Schematic delineating the workflow of Deep Kernel Learning (DKL). Inputs (Red) are encoded into a low dimensional latent space (yellow/orange) which is the last layer of a fully-connected neural network (blue). A non-parametric (infinite parameters) gaussian process regression model (green) is then used to predict the targets (purple) in the latent space.

We hypothesize that the combination of neural networks and GPs in DKL results in a model that is tailored to the target function considered. By enforcing the target function distribution in the latent space to be smooth using GPs, DKL constructs a latent space that may potentially be better suited for optimization than the latent space of a VAE. The smoothness of the target function in the latent space should allow optimization algorithms to move more efficiently towards the global minimum of the target function, while the neural network provides a flexible and powerful way to map the inputs to the latent space.

To explore these assertions, here we apply DKL to the cards dataset previously discussed to investigate its latent space construction and compare it to that of VAE. We note that in this particular analysis, we adopt a static setting of the DKL, where both the input dataset and their corresponding target function outputs are assumed to be available beforehand. Although this setting stands in contrast to the typical process optimization scenario, our goal here is to visualize the latent space of DKL on this toy dataset. In addition to predicting/optimizing continuous target functions, DKL can also be used to predict/optimize categorical target functions. Although the predictions of the GP are continuous, the optimization of categorical target functions can be achieved through a clever construction of the target function. For example, in a binary case with two classes, the target function can be set to one for the class that we are trying to optimize and zero for the other class. This way, the (BO) algorithm, when trying to maximize the target function, will select the classes with higher values of the target function. In the case of the cards dataset, the categorical target function corresponds to the suit of the card. By encoding the target function in this way, the BO algorithm is able to select latent variables that correspond to the desired suit, resulting in the generation of cards with the specified suit.

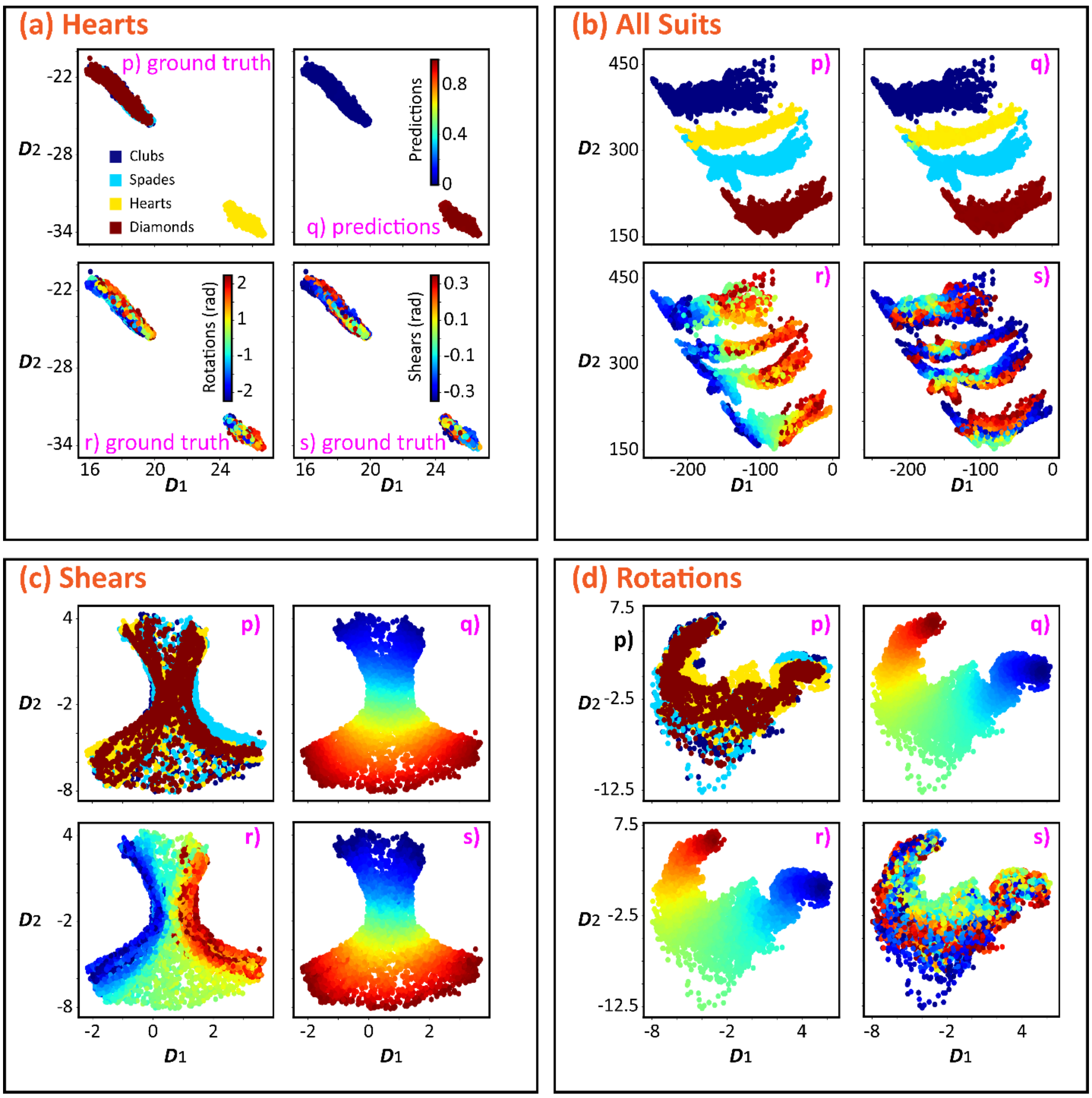

**Figure 3.** Latent space distributions of the DKL on cards dataset when all the input-output pairs are available beforehand. The figure is divided into four parts corresponding to target function optimized in the analysis *viz.*, (a) hearts, (b) all suits, (c) shears, and (d) rotations. Each part is further divided into four subparts where images are represented as the scatter points in the latent space and colored using (p) ground-truth class labels, (q) target function predictions by DKL, (r) ground-truth rotations' distribution, and (s) ground-truth shear distribution. The color bars for the ground-truths are shown in (a) for all the variabilities and since the predictions in the data-rich regime follow the ground truth results closely, the color bars for predictions are omitted for brevity.

The latent dimensions of the DKL will be referred to as $D_1$-$D_2$ to avoid the confusion with the latent dimensions of VAE. The resulting latent distributions of DKL on the cards dataset are shown in Figure 3, where the figure is divided into four parts (a-d) corresponding to four different target functions predicted by the DKL. Each part is further divided into four subparts, where the latent distributions of the DKL are colored with predicted target function (q) and ground truth values of class labels (p), rotations (r), and shear (s). Since the latent distributions of the DKL strongly depend on the target function, we observed four distinct latent distributions for each target function. The first set of results (Figure 3a) corresponds to predicting a categorical target function, where the target function is set to be one for the hearts suit and zero for the other suits. Figures 3ap and 3aq show that the DKL separated the images corresponding to the hearts suit in the latent space. While the rest of the classes are overlapped on top of each other in the latent space, as the target function for the rest of the classes is zero. For the second set of the results, the target function is set to be 0 for clubs, 1 for spades, 2 for hearts and 3 for diamonds. In this scenario, the DKL is forced to separate all the classes. This separation of classes can be observed in Figure 3bp (ground truth) and Figure 3bq (predictions). Only a tiny portion of the hearts class appears to be overlapped with the spades class. This might be because the hearts class at such high rotations and shear are indistinguishable from the spades class. This is the reason why the hearts suit is picked for discussion in the article. However, better-looking results of the other suits can be visualized in the jupyter notebook that accompanies the article. A deeper encoding network can sometimes alleviate this issue. In both cases of the categorical target functions, when the latent distributions are colored using the continuous ground truth variabilities shear (Figures 3as and 3bs) and rotations (Figures 3ar and 3br), they appear non-differentiable in the latent space. This phenomenon of completely ignoring the variabilities that are not part of the target function can also be observed for the categorical variabilities in Figure 3aq, where the classes whose target function values are zeros overlap on top of each other.

The results in Figure 3c and 3d correspond to the cases where the target function is based on continuous variabilities. As with the categorical case, the DKL produces latent spaces that are tailor-made for the target function, with the remaining variabilities in the system ignored. Figures 3c and 3d show the results for the target functions of shears and rotations, respectively. The predictions of the target function in both cases appear to be perfect, as we are operating in a data-rich regime. However, the suit of the image is completely ignored in these cases, as can be observed from Figures 3cp and 3dp. The latent distribution of the rotations while predicting shears (Figure 3cr) appears to be smooth, but this is a serendipitous result and cannot be expected in general. In contrast, for the other case where rotations are part of the target function, the shear distribution (Figure 3ds) appears to be non-differentiable.

Figure 3 presents the latent spaces constructed by DKL, highlighting its ability to create smooth spaces based on target functions while ignoring other variabilities in the system. This contrasts with the latent spaces produced by VAE, which only considers the input dataset and often result in complicated target function distribution in the latent space. The target function distribution in the latent space of the VAE cannot be controlled by the user. The caveat with the latent spaces of DKL arises in the low-data regime, where the target function values are not readily available for all the input data points. Since the DKL's latent space is a function of both input and the value of the target function, only the datapoints whose target function values are available, form the training dataset for DKL. With every additional point, the DKL's latent space evolves to generalize it to the unseen dataset. In contrast, the latent space of the VAE is trained using just the inputs and remains constant throughout the optimization. The capability of the DKL's latent spaces can only be leveraged in the high dimensional optimization when the DKL generalizes the latent space to the complete input dataset based on the few target function values available beforehand. The next setting elucidates the active setting with the DKL where the latent space is trained using only a tiny portion of the dataset. The BO algorithm then picks the next point to be evaluated based on the existing latent space, to optimize the target function.

## 3. Latent distribution for active learning DKL

In this section, we investigate the performance of DKL with BO in the active learning scenario. In this scenario, the DKL has access to the entire input dataset, but the outputs become available sequentially. BO enables the DKL to explore the input space and optimize the target function by balancing the exploitation and exploration of the input space using the acquisition function. Usually, the BO algorithm uses a GP as the surrogate probabilistic model, but in this case, we use the DKL. The DKL's latent space is designed to be tailored to the target function, and the acquisition function exploits this by exploring the latent space based on mean and uncertainty in the prediction. The active learning scenario is useful when evaluating the target function for all input points at once is costly. By training the DKL on a few data points and sequentially optimizing the target function values, the active learning scenario becomes a data-efficient method for optimization of high-dimensional systems. The pseudo-code for BO while using DKL as the surrogate probabilistic model is shown below.

| **Pseudo-code for BO using DKL as the surrogate model** |
| --- |
| **Inputs:** Initial training set, $T = \{x_i, y_i\}_{i=1,\dots,N}$; Array of unmeasured points $\ddot{X}$, Acquisition function *acq*, Number of BO steps $N_{BO}$ |
| |
| for $j$=1, …., $N$ **do** |
| Train a DKL model from scratch using $T$ |
| Compute posterior predictive mean ($\ddot{\mu}$) and covariance ($\ddot{\Sigma}$) on the unmeasured points ($\ddot{X}$) |
| Evaluate acquisition function on the unseen data, $acq_i = \ddot{\mu}_i + \lambda \ddot{\Sigma}_{ii}$ |
| Evaluate the target function ($y_{next}$) at $x_{next} = argmax(acq)$ |
| Update $T$ and $\ddot{X}$ by concatenating $(x_{next}, y_{next})$ to $T$ and deleting $x_{next}$ from $\ddot{X}$ |

After training on a small subset of input-output pairs, the DKL predicts the outputs of the unseen dataset and/or the entire input dataset. The DKL's final layer is a Gaussian process (GP) that produces predictions with mean ($\mu$) and standard deviation ($\sigma$), which quantifies uncertainty in the prediction. An acquisition function is constructed using the predicted mean and standard deviation, evaluated on the input dataset, and the input with the maximum value of the acquisition function in the unseen dataset is chosen for evaluation subsequently. This new evaluated datapoint by an experiment or a simulation is then added to the training set of the DKL, and the process is repeated for a set number of times. The standard deviation enables the DKL to explore unexplored areas, while the mean allows it to explore points with high target function values. For the cards dataset, the DKL is initially trained with the target function outputs of 100 randomly selected input images for each target function, and then sequentially explores 500 datapoints that optimize the target function. The acquisition function choses for the cards dataset is $\mu + 10\sigma$ to balance the exploitation-exploration criterion of the BO algorithm.

In this section, we present the results of the active exploration of DKL on the cards dataset in Figure 4. The figure is divided into two parts: Figure 4a corresponds to optimizing a categorical target function for cards suit, and Figure 4b corresponds to optimizing a continuous target function rotations involved in the image. Each part is further divided into six subparts (p-u) corresponding to latent distributions colored with various properties. The plots in (p-r) are the latent distributions colored with predicted target function values, while the plots (s-u) are colored with ground truth values of class labels, shear, and rotations respectively. The 600 points explored at the end of this analysis including the initial 100 points are plotted in (p), the remaining unexplored points are plotted in (q), and all the input points are plotted in (r-u).

The results of the exploration of the input space for the categorical target function (hearts suit) are shown in Figure 4a. It should be noted that the BO algorithm tries to find the images that maximize the target function while exploring the latent space. The explored points indicate that the majority of images belong to hearts, while a few images of other suits are explored to minimize uncertainty in the latent space. As noted earlier, at the end of the exploration, the DKL only has access to 600 input-output pairs out of 8000 images dataset. The algorithm generalizes the learned patterns to the unseen dataset, as observed from the unexplored points and the whole input dataset plots in Figures 4aq, 4ar, and 4as. However, the generalization is not perfect which can be observed from the thin branch of the hearts suit in Figure 4as that runs through the collection of other suits. The predicted target function values for this thin branch at the end of the exploration are zeros and can be seen in Figure 4ap. More exploration points are necessary to generalize the latent space to all input images. The rest of the suits did not have this issue, and 600 explored points were enough to generalize the latent space to unseen images. The continuous ground truth values in the active-DKL setting also form a non-differentiable distribution in the latent space as they are not part of the target function.

The results of the exploration of the input space for rotations as the target function are shown in Figure 4b. The results here form a similar trend to the categorical target function discussed in Figure 4a. The algorithm explores a majority of the points where the target function has high values, as seen in Figure 4bp. The points near the global minimum are also explored during the process as the uncertainty in predictions near the optima is usually high for GP. The ground truth for this target function is shown in Figure 4bu and the latent space is segregated based on the value of the target function. The algorithm optimally generalizes the latent space to the unseen datapoints, as observed in Figures 4bq and 4br. The other variabilities form a non-differentiable distribution in the latent space, as seen in Figure 4bs for class labels and Figure 4bt for shear. Overall, the DKL algorithm in the active setting does an optimal job in generalizing the learned patterns to the majority of unseen dataset. It should be noted that the active setting of the DKL only has the access to a mere 600 datapoints out of the 8,000 datapoints in the input dataset. With these explanations on the workings of DKL-BO algorithm (also referred to as active setting), we now proceed to deploy it for the application of process optimization in the high-dimensional space.

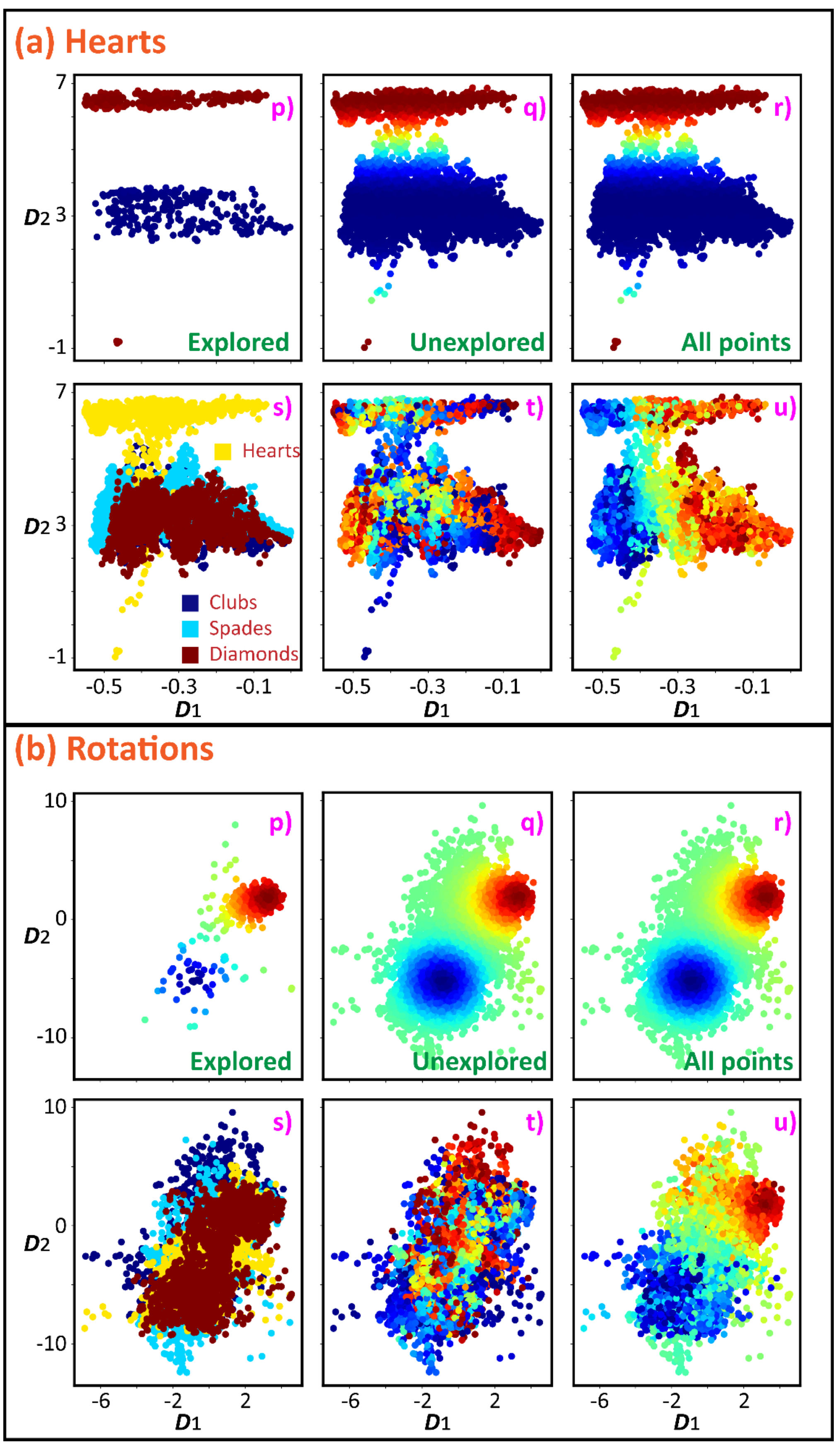

**Figure 4.** Latent space visualization of the DKL on cards dataset when 100 input-output pairs are provided at the initialization step and the BO is allowed to explore for 500 more datapoints while optimizing the target function in the latent space. The target functions considered are (a) categorical hearts suit and (b) continuous rotations. Each part is further divided into six subplots where the input images are embedded using the encoding layer of DKL after the exploration and are colored using (p-r) GP predictions in the latent space, ground-truth (s) class labels, (t) shears' distribution, and (u) rotations' distribution. The scatter points correspond to (p) 600 explored points (q) unexplored points and (r-u) the whole input dataset. The color bars are omitted for brevity.

## 4. Process optimization

The application of methods such as Bayesian Optimization is inherently limited by the high dimensionality of the search space.[31] For example, applications of BO for battery charging are typically based on the representation of current and voltage profiles as a small number of discrete segments or parameters of which yield low dimensional optimization space.[29, 61]

Recently, a number of groups have proposed an approach based on combining the variational autoencoders subsequent Bayesian optimization.[31, 52, 53] In this approach, the family of the domain-specific inputs are encoded to the low-dimensional latent space, and Bayesian Optimization is performed over resultant latent space. Once the maximum is discovered, the latent vector can be decoded to yield optimal trajectory. However, the structure of the latent space in this approach is determined solely by the input data set only. Correspondingly, depending on the distribution of close to optimal and very suboptimal trajectory, the part of the data space occupied by the trajectories of interest and the topology of the corresponding quality surface can be very complex, hindering the BO method.[31]

For the purpose of our study, we focus on analyzing the domain structure evolution in ferroelectric material. To perform the analysis, we use a kinetic lattice model of a ferroelectric system called FerroSIM, which represents a discrete lattice of continuous spins. Although our specific example is based on FerroSIM, the proposed approach can be applied to other systems. FerroSIM is a lattice model that represents a ferroelectric material, first introduced by Ricinschi *et al*.[62] In this model, the order parameter, which is the polarization in the case of a ferroelectric material, is represented as a continuous quantity at each lattice site. The local free energy function is assumed to be of the Ginzburg-Landau-Devonshire (GLD) form, and the polarization at each time step is updated by calculating the derivatives of the total free energy with respect to the polarization field at the previous time step. The aim of this study is to optimize the external electric field curve in time to maximize the target function at the end of the simulation. Three different target functions are chosen for this study *viz*., curl, normalized curl, and the total polarization. The curl at the end of the simulation is defined as the sum of the absolute value of curls at each lattice site, which provides a measure of the domain structure complexity and plays a key role in the material properties. This will be referred to as 'curl' for brevity. At the end of each simulation, the polarization vectors at each site are normalized and the curl is recalculated. This property will be referred to as the normalized curl and gives us insights about the rotation of the polarization field at each site without the effect of magnitude of polarization vectors. The final property is the total polarization and is the magnitude of vector sum of the total polarization in $x$ and $y$ directions. By simulating the polarization dynamics under the influence of an external electric field, we can explore the input space of electric field curves and search for the one that maximizes the target function. The details of the FerroSIM are thoroughly discussed in our previous articles.[50, 63]

As an initial step of the analysis, a domain expert selects the input set of curves deemed relevant. For the FerroSIM case, 7,500 curves of the form $A exp(\alpha t) sin(\omega t) + B$ were chosen by varying the parameters [$A$, $\alpha$, $\omega$, and $B$] within specific intervals.[50] Along with the high dimensional input space which is equal to the sampling frequency of the applied electric field, the challenge of optimizing arises from the complexity of the FerroSIM model and the non-linearity of the considered target functions. Traditional optimization methods may not be suitable for such high-dimensional problems, where the input space can be extremely large and relationship between the input and the target function is complex. Therefore, we used a VAE-BO architecture to optimize the curl in our previous work. The workings of this method and the detailed explanation of the curl surface in VAE's latent space are discussed in ref.[50]

Briefly, the results of a typical FerroSIM simulation are shown in Figure 5a for a sinusoidal applied external electric field. The curl distribution in the latent space of the VAE trained on the 7,500 curves considered is shown in Figure 5b. The hysteresis curve shown in Figure 5a follows the very known trends of the ferroelectric material in the presence of applied external electric field like polarization saturation, remnant polarization, and coercive field. Polarization snapshots at four-different points along the hysteresis curve are shown around the Figure 5a, while these selected points are shown on the hysteresis curve using 'red stars.' Specifically, these points correspond to saturated polarization in $-x$ and $-y$ directions, remnant polarization along the positive polarization-axis, and the coercive field along the positive field-axis. In these snapshots, the polarization vector at each lattice site is shown as an arrow relative to the maximum polarization at each site. Additionally, each lattice site is colored using the vector magnitude of polarization in $x$-direction, as the external electric field is applied only in $x$-direction and held at zero in the $y$-direction.

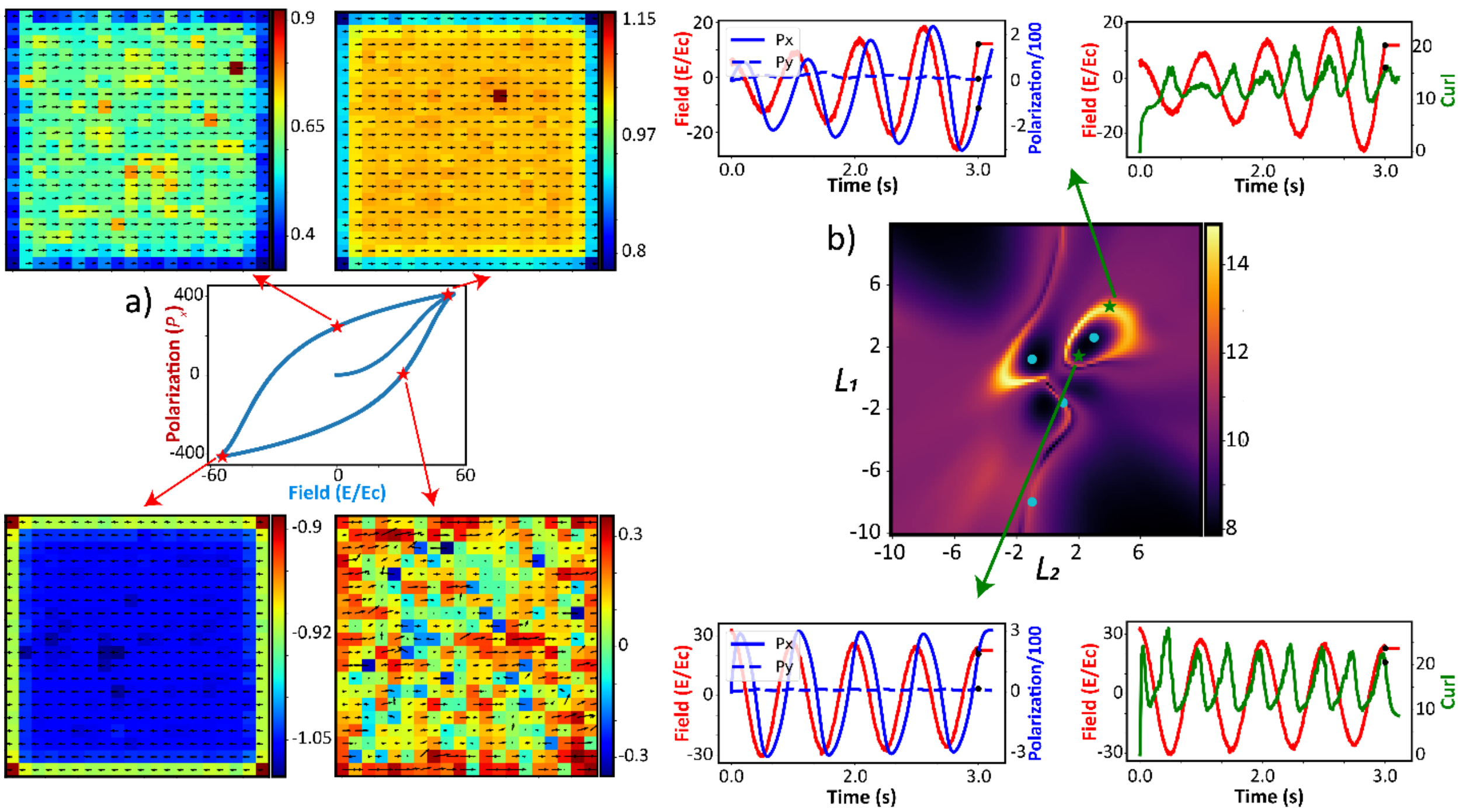

**Figure 5.** (a) Hysteresis loop of a FerroSIM simulation under a sinusoidal external applied field in the x direction. Snapshots of the polarization vectors at saturation field in both positive and negative *x*-direction, coercive field, and remnant polarization. The polarization plots are colored using the *x*-component of the polarization vector. (b) Final value of curl as a function of VAE's latent space. Applied electric field, polarization, and curl *vs*. time of two typical points in the surface are shown around the curl surface. Four other circular scatter points are marked in the latent space that aid the explanation of the shortcomings of the VAE's latent space.

Additionally, the ground-truth curl surface in the latent space of the VAE from our previous work is shown in Figure 5b. The (curl, field) *vs*. time and (polarization, field) *vs*. time plots are shown for two points in the latent space marked with green stars in the curl surface. These points represent two distinct applied electric fields that result in a high and a low value of the final curl. The detailed explanations of the characteristics of the electric fields that result in higher or lower values of curl are found in ref.[50] The limitations of the VAE-BO approach can be immediately understood from the ground truth image. Here, the contrast corresponds to the distribution of the curl in the latent space of the system. Note that the trajectories corresponding to the maximum curl form a manifold in the latent space. However, this manifold has a very complex structure. For example, for point (-0.9, -8) the maxima are aligned on a very narrow ridge separated by the throve from the part of the manifold containing point (1.1, -1.5). Correspondingly, gradient descent or very greedy BO algorithm will fail to traverse from between these two points. Similarly, maxima centered around (3, 2.7) and (-0.9, 1.3) (centers of bright circles) have complex ring-like structures and occupy relatively small part of the latent space. The four points discussed here are marked as circular scatter points in the VAE latent space in Figure 5b. Correspondingly, any GP with a simple kernel will proceed to generate very small kernel length to accommodate the narrow maxima. This in turn necessitates the large number of the GP iterations. We note that this problem, *i.e.*, the properties of the latent manifold containing the behaviors of interest, is extremely well recognized in the ML community.[31, 64] Ultimately, the efficiency of the unsupervised and semi-supervised learning methods, as well as active learning, hinge on how well the manifold is defined and how smooth it is.

## 5. Static DKL on FerroSIM

In this section, we explore the latent spaces of the DKL for the FerroSIM target function when all the outputs are available beforehand. As discussed earlier three different target functions are used to train three DKL networks and the results are shown in Figure 6. Each row in the figure corresponds to the DKL network trained using (a) curl, (b) normalized curl, and (c) total polarization. Each of the latent distributions are colored using the ground truth values of (p) curl, (q) normalized curl, and (r) total polarization. When the data is completely available beforehand, the predictions of the DKL closely follow the ground truth of the target function. So, the diagonal plots which are also marked with 'green stars' in the figure also represent predicted targets of the DKL network.

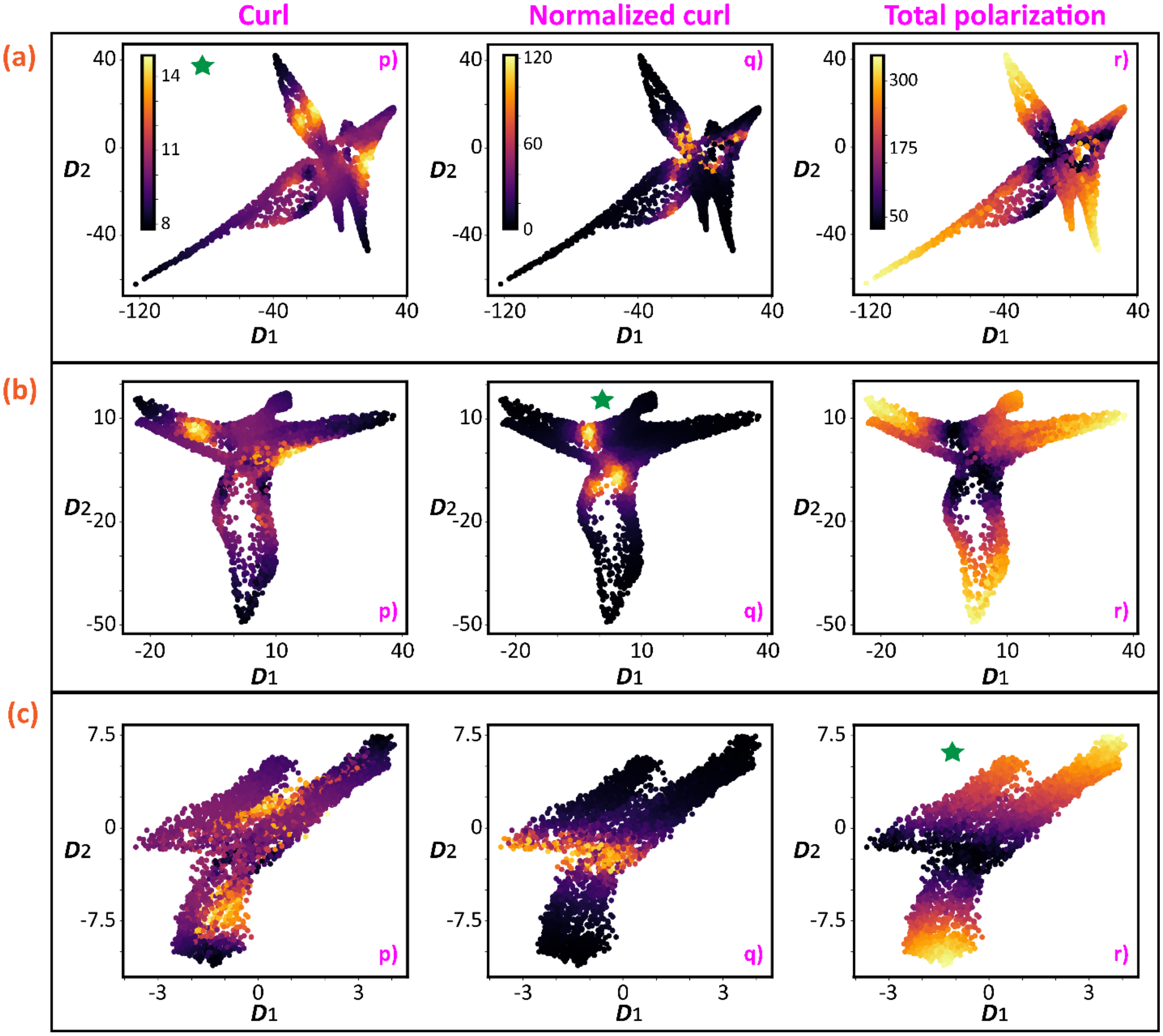

**Figure 6.** Latent space visualization of the DKL on the FerroSIM simulation when all the input-output pairs are available beforehand. Each row (a-c) corresponds to a DKL network trained to predict the final values of (a) curl, (b) normalized curl, and (c) polarization. Each row is further divided into 3 subparts (p-r) where the latent embeddings are colored using the predictions of GP (green-starred plots) and the ground-truth values of other properties (non-starred). The property used to color the scatter points is shown at the top of the column. The color bars for the three properties considered are shown in (a).

For all three properties, the target function in the latent space forms a smooth distribution with 2-3 well connected clusters of the maximum target function values. For example, curl and normalized curl distribution in Figures 6ap and 6bq has two distinct clusters where the target function assumes a high value. For the total polarization case shown in Figure 6cr, the latent distribution has three distinct clusters with high values of the target function. A majority of the curves in input set correspond to a high value of the total polarization. In such scenarios, the VAE's latent space is hard to navigate with multiple local optima, and the BO needs a large number of points before the latent space is fully explored. On the contrary, the DKL does a better job of forming clusters of the input points with optimized values of the target function. The DKL achieves this by clustering the input curves with similar features as well as similar target function properties. This latent space is easier to navigate by the BO algorithm as there are only three different clusters to be explored to optimize the target property. The other ground truth properties do not form a differentiable distribution in the latent space. For example, the normalized curl latent distribution shown in Figure 6aq has points with an overlap between higher and lower values of normalized curl. The same argument is true for the curl distribution in Figure 6cp. As discussed in the case of cards any property that is not directly involved in the construction of the target function will often form a non-differentiable distribution in the latent space.

## 6. Active learning DKL on processes

Finally, we proceed to implement the DKL in the active learning mode for the process optimization. Here, the DKL is provided with the outputs of 100 randomly sampled electric fields from the input dataset. The acquisition function is set to be $\mu + 10\sigma$, which is the same as the one used of the cards dataset and for the exploration of VAE's latent space in ref.[50] The BO then sequentially evaluates the output of 500 more electric field curves while optimizing the target function. The results of the DKL in the active learning setting for the process optimization problem are shown in Figure 7. The figure is divided into 2 parts corresponding to (a) curl and (b) total polarization as the target functions. Each part is then divided into 5 subparts (p-t), where the (p), (q), (r) correspond to the explored points, unexplored points, and the total dataset in the latent space colored with predicted target function values. The latent distributions of the entire dataset colored with ground truth values of the properties that are not part of the target function are shown in (s) and (t).

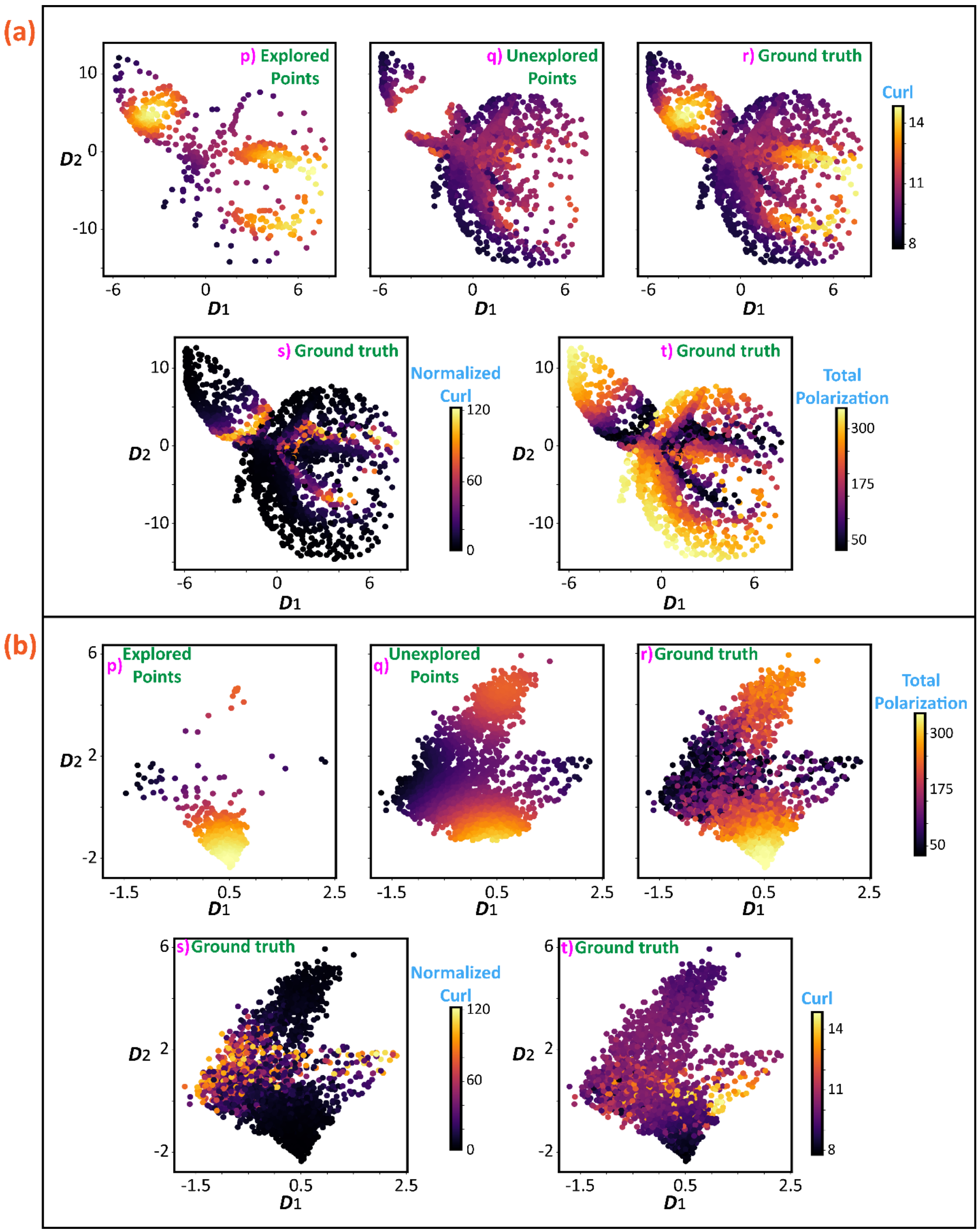

**Figure 7.** Results of the DKL on FerroSIM simulations in the active setting when 100 randomly input-output pairs are provided to the network at the initialization step and the BO is allowed to explore 500 points subsequently. The results are divided into two parts corresponding to the optimized target function (a) curl and (b) total polarization. Each part is further divided into five subparts where the points in the latent space are colored using (p-q) GP predictions of the target function and (r-t) other ground truth properties. (p) Explored points by the network, (q) unexplored points, and (r) all points in the input dataset.

The DKL-BO (active setting) results for the optimization of target function curl are shown in Figure 7a. The target function distribution in this case (Figure 7ar) has three clusters corresponding to the high value of curl. Figures 7ap and 7aq show the explored and unexplored points in the latent space respectively, colored with the predicted value of the curl. These plots show that most of the input curves with high values of the target function are explored by the algorithm. The distributions of other properties considered *viz.*, normalized curl in Figure 7as and total polarization in Figure 7at show that the distributions of these properties are not conducive for optimization in the latent spaces. For example, the polarization distribution (Figure 7at) has multiple local minima, and the normalized curl (Figure 7ats) distribution is non-differentiable.

The results when the total polarization is set as the target function for the algorithm are shown in Figure 7b. Majority of the curves in the input space correspond to a high value of the polarization which complicates the latent space of VAE for optimization purposes. The target function distribution in Figure 7br shows that the DKL is able to cluster these high values of polarization into two clusters. The explored points plot shows that the algorithm completely explored one cluster and even recognized the presence of a second cluster with high values of polarization. This can be deduced from the fact that a few data points from the second cluster are sampled by the BO during the exploration. The cluster that is fully explored by the algorithm comprises the data points with higher values of the polarization which is the reason why the BO has not explored the second cluster within the 600-exploration points limit. The ground truth distribution of the other two properties *viz.*, normalized curl in Figure 7bs and curl in Figure 7bt show that these distributions are non-differentiable *i.e.*, the data points with both high and low values of these properties are overlapped in the latent space.

In Figure 8, we present a comparison of the latent spaces of VAE and DKL for the target functions of curl and total polarization. The distribution of the ground truth target functions in the VAE's latent space is shown in Figure 8a for curl and Figure 8d for total polarization. The smoothness of the VAE's latent spaces for both target functions illustrates the differentiable relationship between the inputs and target functions. However, these latent spaces are found to be complicated with complex local optima for curl and a large number of local optima for total polarization. As a result, any greedy optimization algorithm gets trapped in these local optima, or in the case of BO, a large number of exploration points are necessary to model the kernel length and explore all the local optima. The results of exploration of curl in the latent space using BO is thoroughly discussed in discussed in ref.[50] The DKL latent distributions for the static setting of both target functions are shown in Figure 8b (curl) and 8e (polarization). These latent spaces in the data-rich regime do not provide insights into the optimization process. The latent distributions of the ground-truth target functions in the latent space at the end of the exploration (600 data points) by DKL-BO (active setting) are shown in Figures 8c and 8f for curl and total polarization, respectively. These low-dimensional latent spaces are much more conducive to the navigation of optimization algorithms, with fewer well-connected local optima.

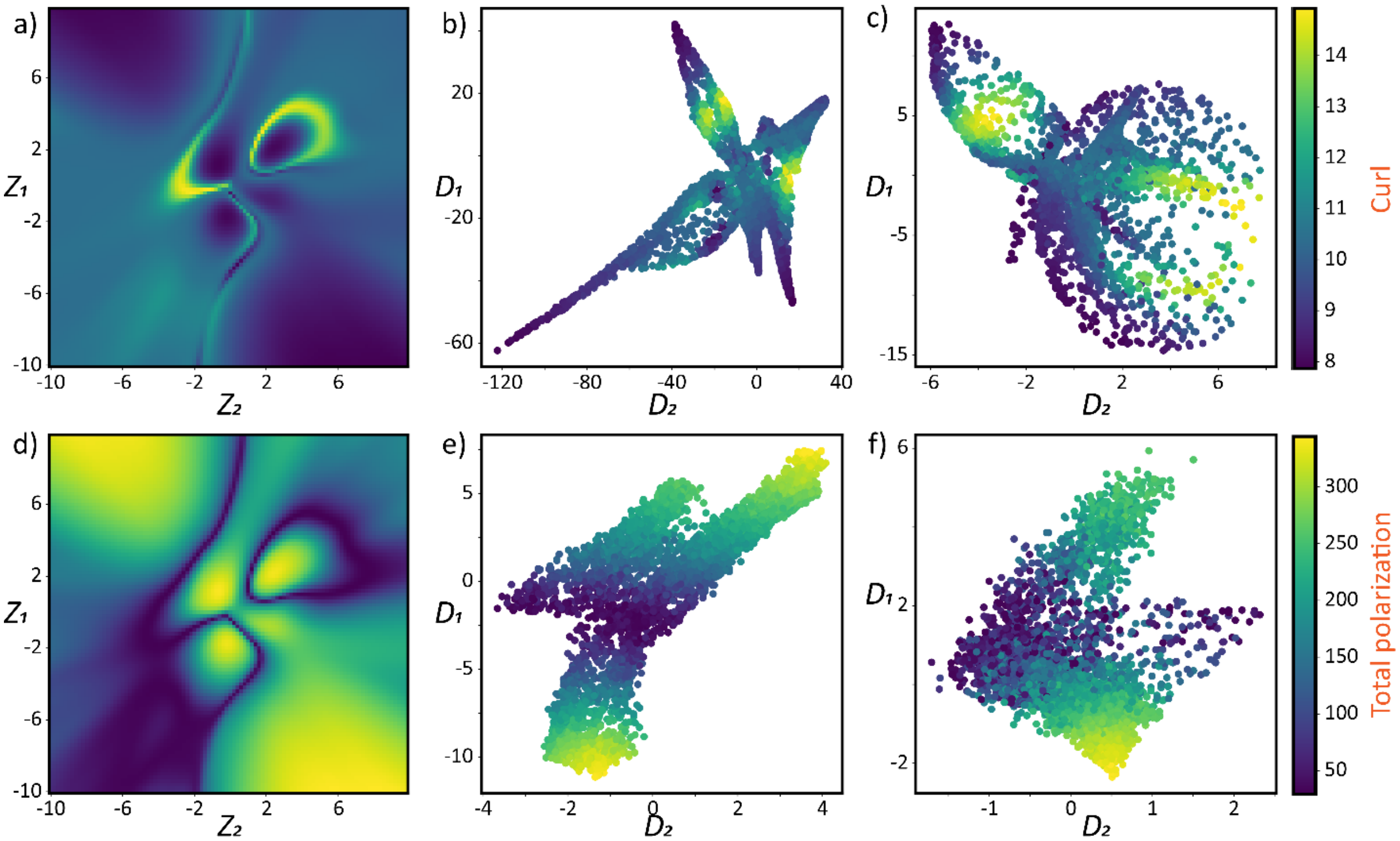

**Figure 8.** The ground-truth distribution of (a) curl and (d) total polarization in the latent space of VAE trained on the entire input dataset. Latent space distribution of the static-setting DKL colored with the ground-truth values of the target function (b) curl and (e) total polarization. Latent distributions at the end of the exploration (600 datapoints) of the active-setting DKL colored with ground truth values of the target function (c) curl and (f) total polarization.

## 7. Summary

We have compared the process optimization via BO optimization in the latent space of the VAE-encoded process trajectories, deep kernel learning over the static data set (fully known output data) and active DKL learning with partial data. The structure of the latent data space for VAE is determined solely by the properties of the input data set and resulting functionality often has very complex structure in the latent space. The latter in turn precludes optimization *via.*, gradient descent or greedy Bayesian Optimization and imposes the significant limitations on the upper bound of the kernel length in GP. This in turn necessitates a large number of the learning steps in the case of BO to explore the latent space well.

The DKL on the full data set (known ground truth) forms much more compact distributions of the functionality of interest. In this case, the latent encoding reflects both the properties of the data and the properties of the target set. The shape of the latent distribution can be complex, but the functionality of interest forms much better formed regions. Correspondingly, the alternative functionalities tend to be more abrupt. We note that this case corresponds to the known ground truth and shown here for illustration.

Finally, the active learning DKL forms well behaved latent distributions. The functionality of interest tends to form smaller number of broad maxima in the latent space. For the scenarios analyzed in this article, the distributions look smooth, and we pose that they can be potentially differentiable due to the presence of GP in DKL. At the same time, counterfactual parameters show much more discontinuous and non-differentiable distributions.

We pose that this finding can offer major implications for optimization and discovery problems over the high-dimensional and potentially nondifferentiable spaces, such as natural language processing or molecular and material optimization. In this case, the properties (smoothness and localization) of the manifold play the central role in the discovery process, and active learning DKL shows clear advantage compared to the embedding formed based on data only.

**Data availability statement:**

- The complete analysis DKL, VAE, and rVAE on the cards dataset can be visualized and reproduced using the jupyter notebook at https://github.com/saimani5/Notebooks_for_papers/blob/main/dkl_bo_cards.ipynb
- The complete analysis of DKL on the FerroSIM can be visualized and reproduced using the jupyter notebook at https://github.com/saimani5/Notebooks_for_papers/blob/main/ferrosim_dkl.ipynb
- VAE and rVAE use a package 'atomai' based on pyTorch: https://github.com/pycroscopy/atomai
- DKL uses a package 'gpax' built on top of NumPyro, JAX, and Haiku: https://github.com/ziatdinovmax/gpax
- GitHub repository of FerroSIM: https://github.com/ramav87/FerroSim

**Acknowledgements:**

This work was supported by the Energy Frontier Research Centers program: CSSAS–The Center for the Science of Synthesis Across Scales–under Award Number DE-SC0019288, located at the University of Washington, and the modeling and process optimization by the US Department of Energy Office of Science under the Materials Sciences and Engineering Division of the Basic Energy Sciences program. The Bayesian optimization and Deep Kernel Learning research was supported by the Center for Nanophase Materials Sciences (CNMS) which is a US Department of Energy, Office of Science User Facility at Oak Ridge National Laboratory.

## References

1. Ghosh, A.; Sumpter, B. G.; Dyck, O.; Kalinin, S. V.; Ziatdinov, M., Ensemble learning-iterative training machine learning for uncertainty quantification and automated experiment in atom-resolved microscopy. *npj Computational Materials* **2021,** *7* (1), 100.
2. Wang, P.; Chen, P.; Yuan, Y.; Liu, D.; Huang, Z.; Hou, X.; Cottrell, G. In *Understanding Convolution for Semantic Segmentation*, 2018 IEEE Winter Conference on Applications of Computer Vision (WACV), 12-15 March 2018; 2018; pp 1451-1460.
3. Gupta, S.; Arbeláez, P.; Girshick, R.; Malik, J., Indoor Scene Understanding with RGB-D Images: Bottom-up Segmentation, Object Detection and Semantic Segmentation. *International Journal of Computer Vision* **2015,** *112* (2), 133-149.
4. Lewis-Beck, C.; Lewis-Beck, M., *Applied Regression: An Introduction*. SAGE Publications: 2015.
5. Pinson, P.; Han, L.; Kazempour, J., Regression markets and application to energy forecasting. *Top* **2022,** *30* (3), 533-573.
6. Chen, W.; Xiao, D.; Li, X., Classification, application, and creation of landscape indices. *Ying Yong Sheng tai xue bao= The Journal of Applied Ecology* **2002,** *13* (1), 121-125.
7. Singhal, G.; Bansod, B.; Mathew, L., Unmanned aerial vehicle classification, applications and challenges: A review. **2018**.
8. Hilas, C. S.; Mastorocostas, P. A., An application of supervised and unsupervised learning approaches to telecommunications fraud detection. *Knowledge-Based Systems* **2008,** *21* (7), 721-726.
9. Celebi, M. E.; Aydin, K., *Unsupervised learning algorithms*. Springer: 2016; Vol. 9.
10. Wei, B.; Ye, X.; Long, C.; Du, Z.; Li, B.; Yin, B.; Yang, X., Discriminative Active Learning for Robotic Grasping in Cluttered Scene. *IEEE Robotics and Automation Letters* **2023**.
11. Moya, B.; Badías, A.; González, D.; Chinesta, F.; Cueto, E., A thermodynamics-informed active learning approach to perception and reasoning about fluids. *Computational Mechanics* **2023**.
12. Behera, S.; Samantaray, B. K.; Nayak, J.; Hota, N., Review of Control Principles for Active Learning in Robotics.
13. Bose, A.; Bruno, J.; Dames, P.; Bai, L. In *Time Constraint Finite-Horizon Path Planning Solution for Micromouse Extreme Problem*, 2022 IEEE 19th International Conference on Mobile Ad Hoc and Smart Systems (MASS), 19-23 Oct. 2022; 2022; pp 325-331.
14. Dhananjaya, M. M.; Kumar, V. R.; Yogamani, S. In *Weather and Light Level Classification for Autonomous Driving: Dataset, Baseline and Active Learning*, 2021 IEEE International Intelligent Transportation Systems Conference (ITSC), 19-22 Sept. 2021; 2021; pp 2816-2821.
15. Liang, Z.; Xu, X.; Deng, S.; Cai, L.; Jiang, T.; Jia, K. Exploring Diversity-based Active Learning for 3D Object Detection in Autonomous Driving 2022, p. arXiv:2205.07708. https://ui.adsabs.harvard.edu/abs/2022arXiv220507708L (accessed May 01, 2022).
16. Feng, D.; Wei, X.; Rosenbaum, L.; Maki, A.; Dietmayer, K. In *Deep Active Learning for Efficient Training of a LiDAR 3D Object Detector*, 2019 IEEE Intelligent Vehicles Symposium (IV), 9-12 June 2019; 2019; pp 667-674.
17. Denzler, P.; Ziegler, M.; Jacobs, A.; Eiselein, V.; Neumaier, P.; Köppel, M. In *Multi-Sensor Data Annotation Using Sequence-based Active Learning*, 2022 IEEE/RSJ International Conference on Intelligent Robots and Systems (IROS), 23-27 Oct. 2022; 2022; pp 258-263.
18. Schneegans, J.; Bieshaar, M.; Sick, B., A Practical Evaluation of Active Learning Approaches for Object Detection. **2022**.
19. Kusne, A. G.; Yu, H.; Wu, C.; Zhang, H.; Hattrick-Simpers, J.; DeCost, B.; Sarker, S.; Oses, C.; Toher, C.; Curtarolo, S.; Davydov, A. V.; Agarwal, R.; Bendersky, L. A.; Li, M.;

Mehta, A.; Takeuchi, I., On-the-fly closed-loop materials discovery via Bayesian active learning. *Nature Communications* **2020,** *11* (1), 5966.
20. Lookman, T.; Balachandran, P. V.; Xue, D.; Yuan, R., Active learning in materials science with emphasis on adaptive sampling using uncertainties for targeted design. *npj Computational Materials* **2019,** *5* (1), 21.
21. Ziatdinov, M. A.; Liu, Y.; Morozovska, A. N.; Eliseev, E. A.; Zhang, X.; Takeuchi, I.; Kalinin, S. V., Hypothesis Learning in Automated Experiment: Application to Combinatorial Materials Libraries. *Advanced Materials* **2022,** *34* (20), 2201345.
22. Borg, C. K. H.; Muckley, E. S.; Nyby, C.; Saal, J. E.; Ward, L.; Mehta, A.; Meredig, B., Quantifying the performance of machine learning models in materials discovery. *Digital Discovery* **2023**.
23. Yuan, X.; Zhou, Y.; Peng, Q.; Yang, Y.; Li, Y.; Wen, X., Active learning to overcome exponential-wall problem for effective structure prediction of chemical-disordered materials. *npj Computational Materials* **2023,** *9* (1), 12.
24. Ziatdinov, M.; Liu, Y.; Kalinin, S. V. Active learning in open experimental environments: selecting the right information channel(s) based on predictability in deep kernel learning 2022, p. arXiv:2203.10181. https://ui.adsabs.harvard.edu/abs/2022arXiv220310181Z (accessed March 01, 2022).
25. Ghosh, A.; Kalinin, S.; Ziatdinov, M., Hypothesis-driven active learning over the chemical space. *Bulletin of the American Physical Society* **2023**.
26. AlFaraj, Y.; Mohapatra, S.; Shieh, P.; Husted, K.; Ivanoff, D.; Lloyd, E.; Cooper, J.; Dai, Y.; Singhal, A.; Moore, J., A Model Ensemble Approach Enables Data-Driven Property Prediction for Chemically Deconstructable Thermosets in the Low Data Regime. **2023**.
27. Griffiths, R.-R.; Klarner, L.; Ravuri, A.; Truong, S.; Rankovic, B.; Du, Y.; Jamasb, A.; Schwartz, J.; Tripp, A.; Kell, G. In *GAUCHE: A library for Gaussian processes and Bayesian optimisation in chemistry*, presented at the ICML 2022 Workshop on Adaptive Experimental Design and Active Learning in the Real World (ReALML 2022), Baltimore: 2022.
28. Chen, H.; Li, X. R. In *Distributed active learning with application to battery health management*, 14th International Conference on Information Fusion, 5-8 July 2011; 2011; pp 1-7.
29. Attia, P. M.; Grover, A.; Jin, N.; Severson, K. A.; Markov, T. M.; Liao, Y.-H.; Chen, M. H.; Cheong, B.; Perkins, N.; Yang, Z.; Herring, P. K.; Aykol, M.; Harris, S. J.; Braatz, R. D.; Ermon, S.; Chueh, W. C., Closed-loop optimization of fast-charging protocols for batteries with machine learning. *Nature* **2020,** *578* (7795), 397-402.
30. Suthar, B.; Ramadesigan, V.; De, S.; Braatz, R. D.; Subramanian, V. R., Optimal charging profiles for mechanically constrained lithium-ion batteries. *Physical Chemistry Chemical Physics* **2014,** *16* (1), 277-287.
31. Siivola, E.; Paleyes, A.; González, J.; Vehtari, A., Good practices for Bayesian optimization of high dimensional structured spaces. *Applied AI Letters* **2021,** *2* (2), e24.
32. Sohn, K.; Lee, H.; Yan, X., Learning structured output representation using deep conditional generative models. *Advances in neural information processing systems* **2015,** *28*.
33. Bepler, T.; Zhong, E.; Kelley, K.; Brignole, E.; Berger, B., Explicitly disentangling image content from translation and rotation with spatial-VAE. *Advances in Neural Information Processing Systems* **2019,** *32*.
34. Kingma, D. P.; Welling, M., An introduction to variational autoencoders. *Foundations and Trends® in Machine Learning* **2019,** *12* (4), 307-392.
35. Guo, Y.; Liao, W.; Wang, Q.; Yu, L.; Ji, T.; Li, P., Multidimensional Time Series Anomaly Detection: A GRU-based Gaussian Mixture Variational Autoencoder Approach. In *Proceedings of The 10th Asian Conference on Machine Learning*, Jun, Z.; Ichiro, T., Eds. PMLR: Proceedings of Machine Learning Research, 2018; Vol. 95, pp 97--112.

36. Chen, T.; Liu, X.; Xia, B.; Wang, W.; Lai, Y., Unsupervised anomaly detection of industrial robots using sliding-window convolutional variational autoencoder. *IEEE Access* **2020,** *8*, 47072-47081.
37. Suh, S.; Chae, D. H.; Kang, H. G.; Choi, S. In *Echo-state conditional variational autoencoder for anomaly detection*, 2016 International Joint Conference on Neural Networks (IJCNN), 24-29 July 2016; 2016; pp 1015-1022.
38. Pu, Y.; Gan, Z.; Henao, R.; Yuan, X.; Li, C.; Stevens, A.; Carin, L., Variational autoencoder for deep learning of images, labels and captions. *Advances in neural information processing systems* **2016,** *29*.
39. Walker, J.; Doersch, C.; Gupta, A.; Hebert, M. In *An Uncertain Future: Forecasting from Static Images Using Variational Autoencoders*, Computer Vision – ECCV 2016, Cham, 2016//; Leibe, B.; Matas, J.; Sebe, N.; Welling, M., Eds. Springer International Publishing: Cham, 2016; pp 835-851.
40. Shao, H.; Yao, S.; Sun, D.; Zhang, A.; Liu, S.; Liu, D.; Wang, J.; Abdelzaher, T. In *Controlvae: Controllable variational autoencoder*, International Conference on Machine Learning, PMLR: 2020; pp 8655-8664.
41. Hou, X.; Shen, L.; Sun, K.; Qiu, G. In *Deep feature consistent variational autoencoder*, 2017 IEEE winter conference on applications of computer vision (WACV), IEEE: 2017; pp 1133-1141.
42. Liu, Z.-S.; Siu, W.-C.; Wang, L.-W.; Li, C.-T.; Cani, M.-P. In *Unsupervised real image super-resolution via generative variational autoencoder*, Proceedings of the IEEE/CVF conference on computer vision and pattern recognition workshops, 2020; pp 442-443.
43. Du, Y.; Xu, J.; Qiu, Q.; Zhen, X.; Zhang, L. In *Variational image deraining*, Proceedings of the IEEE/CVF Winter Conference on Applications of Computer Vision, 2020; pp 2406-2415.
44. Zhou, L.; Cai, C.; Gao, Y.; Su, S.; Wu, J. In *Variational autoencoder for low bit-rate image compression*, Proceedings of the IEEE Conference on Computer Vision and Pattern Recognition Workshops, 2018; pp 2617-2620.
45. Alves de Oliveira, V.; Chabert, M.; Oberlin, T.; Poulliat, C.; Bruno, M.; Latry, C.; Carlavan, M.; Henrot, S.; Falzon, F.; Camarero, R., Reduced-complexity end-to-end variational autoencoder for on board satellite image compression. *Remote Sensing* **2021,** *13* (3), 447.
46. Ignatans, R.; Ziatdinov, M.; Vasudevan, R.; Valleti, M.; Tileli, V.; Kalinin, S. V., Latent Mechanisms of Polarization Switching from In Situ Electron Microscopy Observations. *Advanced Functional Materials* **2022,** *32* (23), 2100271.
47. Valleti, S. M. P.; Kalinin, S. V.; Nelson, C. T.; Peters, J. J. P.; Dong, W.; Beanland, R.; Zhang, X.; Takeuchi, I.; Ziatdinov, M., Unsupervised learning of ferroic variants from atomically resolved STEM images. *AIP Advances* **2022,** *12* (10), 105122.
48. Kalinin, S. V.; Zhang, S.; Valleti, M.; Pyles, H.; Baker, D.; De Yoreo, J. J.; Ziatdinov, M., Disentangling Rotational Dynamics and Ordering Transitions in a System of Self-Organizing Protein Nanorods via Rotationally Invariant Latent Representations. *ACS Nano* **2021,** *15* (4), 6471-6480.
49. Kalinin, S. V.; Dyck, O.; Jesse, S.; Ziatdinov, M., Exploring order parameters and dynamic processes in disordered systems via variational autoencoders. *Science Advances* **2021,** *7* (17), eabd5084.
50. Valleti, M.; Vasudevan, R. K.; Ziatdinov, M. A.; Kalinin, S. V., Bayesian optimization in continuous spaces via virtual process embeddings. *Digital Discovery* **2022,** *1* (6), 910-925.
51. Huang, W.; Zhao, D.; Sun, F.; Liu, H.; Chang, E., Scalable Gaussian process regression using deep neural networks. In *Proceedings of the 24th International Conference on Artificial Intelligence*, AAAI Press: Buenos Aires, Argentina, 2015; pp 3576–3582.
52. Griffiths, R.-R.; Hernández-Lobato, J. M., Constrained Bayesian optimization for automatic chemical design using variational autoencoders. *Chemical Science* **2020,** *11* (2), 577-586.

53. Kusner, M. J.; Paige, B.; Hernández-Lobato, J. M. In *Grammar variational autoencoder*, International conference on machine learning, PMLR: 2017; pp 1945-1954.
54. Wilson, A. G.; Hu, Z.; Salakhutdinov, R.; Xing, E. P., Deep Kernel Learning. In *Proceedings of the 19th International Conference on Artificial Intelligence and Statistics*, Arthur, G.; Christian, C. R., Eds. PMLR: Proceedings of Machine Learning Research, 2016; Vol. 51, pp 370--378.
55. Liu, Y.; Vasudevan, R. K.; Kelley, K. P.; Funakubo, H.; Ziatdinov, M.; Kalinin, S. V., Learning the right channel in multimodal imaging: automated experiment in piezoresponse force microscopy. *npj Computational Materials* **2023,** *9* (1), 34.
56. Ziatdinov, M.; Liu, Y.; Kelley, K.; Vasudevan, R.; Kalinin, S. V., Bayesian Active Learning for Scanning Probe Microscopy: From Gaussian Processes to Hypothesis Learning. *ACS Nano* **2022,** *16* (9), 13492-13512.
57. Keith, J. A.; Vassilev-Galindo, V.; Cheng, B.; Chmiela, S.; Gastegger, M.; Müller, K.-R.; Tkatchenko, A., Combining Machine Learning and Computational Chemistry for Predictive Insights Into Chemical Systems. *Chemical Reviews* **2021,** *121* (16), 9816-9872.
58. You, J.; Li, X.; Low, M.; Lobell, D.; Ermon, S., Deep Gaussian Process for Crop Yield Prediction Based on Remote Sensing Data. *Proceedings of the AAAI Conference on Artificial Intelligence* **2017,** *31* (1).
59. Garnett, R., *Bayesian Optimization*. Cambridge University Press: 2023.
60. Blei, D. M.; Kucukelbir, A.; McAuliffe, J. D., Variational Inference: A Review for Statisticians. *Journal of the American Statistical Association* **2017,** *112* (518), 859-877.
61. Torchio, M.; Wolff, N. A.; Raimondo, D. M.; Magni, L.; Krewer, U.; Gopaluni, R. B.; Paulson, J. A.; Braatz, R. D. In *Real-time model predictive control for the optimal charging of a lithium-ion battery*, 2015 American Control Conference (ACC), 1-3 July 2015; 2015; pp 4536-4541.
62. Dan, R.; Catalin, H.; Constantin, P.; Liliana, M.; Vasile, T.; Masanori, O., Analysis of ferroelectric switching in finite media as a Landau-type phase transition. *Journal of Physics: Condensed Matter* **1998,** *10* (2), 477.
63. Kalinin, S. V.; Ziatdinov, M.; Vasudevan, R. K., Guided search for desired functional responses via Bayesian optimization of generative model: Hysteresis loop shape engineering in ferroelectrics. *Journal of Applied Physics* **2020,** *128* (2), 024102.
64. Eismann, S.; Levy, D.; Shu, R.; Bartzsch, S.; Ermon, S. In *Bayesian optimization and attribute adjustment*, Conference on Uncertainty in Artificial Intelligence, 2018.